\documentclass{egpubl}
\usepackage{eurovis2026}

\EuroVisGDxDR %

\usepackage[T1]{fontenc}
\usepackage{dfadobe}

\usepackage{cite}  %
\BibtexOrBiblatex
\electronicVersion
\PrintedOrElectronic
\ifpdf \usepackage[pdftex]{graphicx} \pdfcompresslevel=9
\else \usepackage[dvips]{graphicx} \fi

\usepackage{egweblnk}
\usepackage{todonotes}
\usepackage{paralist}
\usepackage{amssymb}
\usepackage{amsmath}

\title[MDS by SGD]{Bridging Graph Drawing and Dimensionality Reduction with Stochastic Stress Optimization}

\author[D. Hangan \& S. Kobourov \& J. Miller]
{\parbox{\textwidth}{\centering D. Hangan, %
        S. Kobourov,\orcid{0000-0002-0477-2724}
        and J. Miller\orcid{0000-0002-0567-785X} 
        }
        \\
{\parbox{\textwidth}{\centering Technical University of Munich
       }
}
}

\begin{document}

\maketitle
\begin{abstract}
    Both Dimensionality Reduction (DR) and Graph Drawing (GD) aim to visualize abstract, non-linear structures, yet rely on different optimization paradigms. This contrast is evident in Multidimensional Scaling (MDS), which typically depends on the SMACOF algorithm despite graph drawing results showing that simpler stochastic optimization schemes can be more effective for the same objective.
    We bridge these domains by adapting Stochastic Gradient Descent (SGD) techniques from graph drawing to vector data embedding. We present a scikit-learn–compatible estimator that minimizes global stress through local pairwise updates, improving upon the existing implementation.
    Experiments on standard high-dimensional benchmarks show that our stochastic solver converges substantially faster than SMACOF while achieving comparable or lower stress. 

\end{abstract}

\section{Introduction}

Graph Drawing (GD) and Dimensionality Reduction (DR) pursue the same fundamental objective: embedding complex structures into interpretable low-dimensional spaces for visualization. Yet, despite this conceptual overlap, the communities have evolved largely in isolation, particularly so in recent years.
A notable example is \textit{Multidimensional Scaling} (MDS). While the GD literature demonstrated in 2018 that stress-based layouts can be optimized effectively using stochastic optimization schemes, the DR community has continued to rely on the SMACOF 
~\cite{deleeuw1977,de2009multidimensional} algorithm. 

SMACOF is mathematically elegant
but inherently a batch method: each iteration requires a full evaluation of stress over all pairs before an update to any position is made.
In contrast, stress objectives in GD now largely perform optimization through a particular form of \textit{Stochastic Gradient Descent} (SGD), which measures local pairwise error and adjusts positions immediately~\cite{zheng2018graph}. Despite clear empirical and practical advantages reported in GD~\cite{borsig2020stochastic,giovannangeli2022forbid,ahmed2022multicriteria}, these stochastic advances have not been widely adopted in DR for MDS.

In this work, we bridge the gap 
by transferring SGD-based stress optimization from GD to the objective of metric MDS on vector data, reinterpreting MDS as a stream of local pairwise interactions. This enables stochastic updates that more effectively explore the non-convex stress landscape and often avoids the local minima that can trap deterministic batch solvers.

\begin{figure}
    \centering
    \includegraphics[width=0.9\linewidth]{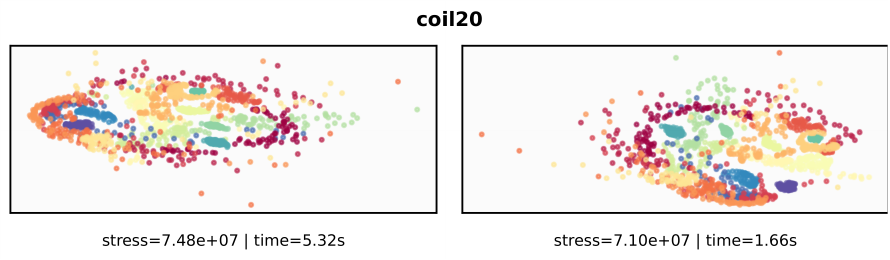}

    \parbox[c]{0.49\linewidth}{\centering SMACOF}
    \parbox[c]{0.49\linewidth}{\centering SGD-MDS}
    \caption{An embedding of coil20 by SMACOF (left) and SGD-MDS (right), which found a lower stress embedding in less time. }
    \label{fig:embedding-compare}
    \vspace{-0.5cm}
\end{figure}

We present 
a robust, scikit-learn compatible \texttt{SGD-MDS} estimator that improves both performance and scalability. Empirical evaluations demonstrate that our solver nearly always converges faster than SMACOF, and reaches comparable or lower stress values; see Figure~\ref{fig:embedding-compare}. 
Although individual epochs are computationally heavier, SGD often requires only a small number of epochs to stabilize, whereas SMACOF commonly needs hundreds of full iterations. SGD also exhibits strong robustness to initialization. 
Finally, we introduce a ``Lazy'' execution mode which computes distances on-the-fly
 which reduces auxiliary memory complexity from $O(N^2)$ to only constant additional space.
 This enables exact MDS
 to scale to datasets with more than 20,000 samples, 
 where standard matrix-based implementations become impractical. Our implementation is available open-source at {\tiny\url{https://github.com/DanielHangan01/scikit-learn}}.

\section{Background and Related Work}

We refer to Metric Multidimensional Scaling (MDS) as the minimization of the stress function:
\begin{equation}
    \label{eq:stress}
    \sigma(X) = \sum_{i=1}^N\sum_{j=i+1}^N w_{ij} \left( \delta_{ij} - d_{ij}(X) \right)^2
\end{equation}
where $\delta_{ij}$ denote the input dissimilarities, $d_{ij}(X)$ are the Euclidean distances in the embedding, and $w_{ij}$ are weights. 
In DR, uniform weights $w_{ij}=1$ are common~\cite{buja2008data} while in GD inverse squared distances $w_{ij} = \delta_{ij}^{-2}$ are used~\cite{gansner2004graph}. Up to normalization constants and weighting choices, the same objective appears in both communities.

In DR, stress-based MDS traces back to Shepard,
Kruskal,
and Sammon
though terminology distinguishes classical,
metric,
and non-metric variants~\cite{borg2005modern}. In GD, the same objective underlies the Kamada–Kawai~\cite{kamada1989algorithm} algorithm and later stress-majorization approaches such as those implemented in \texttt{Neato}~\cite{gansner2004graph}. Despite differing terminology and interpretations, both fields optimize essentially the same stress formulation.

Typically, in DR the dissimilarities $\delta_{ij}$ are derived from feature-based data. However, the stress objective itself is agnostic to the origin of these values.
An early observation in the GD community~\cite{kamada1989algorithm} was that graph-theoretic shortest-path distances can serve directly as input dissimilarities. Under this interpretation, a drawing is considered good if it 
represents the structural information encoded in graph distances. Minimizing stress 
amounts to optimizing how accurately these distances are preserved in the embedding.
Consequently, stress has become one of the most widely used non-aesthetic quality criteria in graph drawing~\cite{di2024evaluating}. Moreover, empirical studies report a strong association between low stress values and visually effective layouts~\cite{mooney2024perception}.

From an optimization perspective, the MDS problem is identical regardless of whether the dissimilarities arise from feature-based data or graph-theoretic distances. The stress objective and its gradients are unchanged; only the numerical values of $\delta_{ij}$ differ. Nevertheless, the DR and GD communities have largely treated these settings as distinct problems. This separation is most apparent in the choice of optimization strategies used to minimize stress.

Historically, the standard optimization approach for MDS is SMACOF~\cite{deleeuw1977,de2009multidimensional,borg2005modern}, a majorization--minimization algorithm that guarantees monotonic stress reduction. SMACOF (\emph{Scaling by MAjorizing a COmplicated Function}) performs global configuration updates via the Guttman transform, aggregating contributions from all $\binom{N}{2}$ pairs at each iteration. Updates are therefore fully synchronous and deterministic, and the algorithm converges to a local minimum determined by the initialization. 
In typical implementations, the full dissimilarity matrix is stored explicitly, resulting in $O(N^2)$ memory complexity. In practice, this quadratic storage requirement limits exact applications to moderate dataset sizes, often only a few thousand points on conventional hardware.

\paragraph{Stochastic Optimization for MDS:}
Prior efforts to apply stochastic optimization to MDS of data are relatively limited. Rajawat and Kumar~\cite{rajawat2017} proposed a ``Stochastic SMACOF'' algorithm, but their work was motivated primarily by online learning in wireless sensor networks where data arrives sequentially.
However, they do not consider static visualization.
In particular, the potential benefits of stochasticity for navigating the non-convex stress landscape in an offline setting, as well as implications for memory scalability, were not examined.
We adapt the work of Zheng et al.~\cite{zheng2018graph} to the DR setting.
They demonstrated that Stochastic Gradient Descent (SGD) can empirically outperform the traditional SMACOF-based optimization for graph drawing.
A key advantage of their approach is immediate feedback: whereas batch solvers aggregate all pairwise contributions before updating the configuration, SGD updates the embedding after processing individual pairs. This enables structural information to propagate continuously during a pass over the data. Combined with the inherent stochasticity of pair sampling, this behavior facilitates broader exploration of the non-convex stress landscape. Zheng et al.'s method has since been incorporated into widely used graph drawing software such as Graphviz (Neato)~\cite{gansner2004graph}, and has been the subject of successful replication studies~\cite{borsig2020stochastic} as well as further extensions~\cite{giovannangeli2022forbid,ahmed2022multicriteria}.

\paragraph{Bridging GD and DR:}
Recent work has advocated a unified perspective on DR and GD. Paulovich et al.~\cite{paulovich2024} present a framework connecting standard DR pipelines with graph drawing procedures, noting that neighborhood construction and layout optimization closely parallel classical GD.
Our work demonstrates a realization
of this connection: we apply the stochastic solver strategy developed in GD to the
stress objective used in DR. The resulting method results in the same embedding quality expected of an established DR algorithm with the empirical scalability and optimization behavior characteristic of stochastic graph drawing algorithms.

\begin{figure*}[h]
    \centering
    \includegraphics[width=0.9\linewidth]{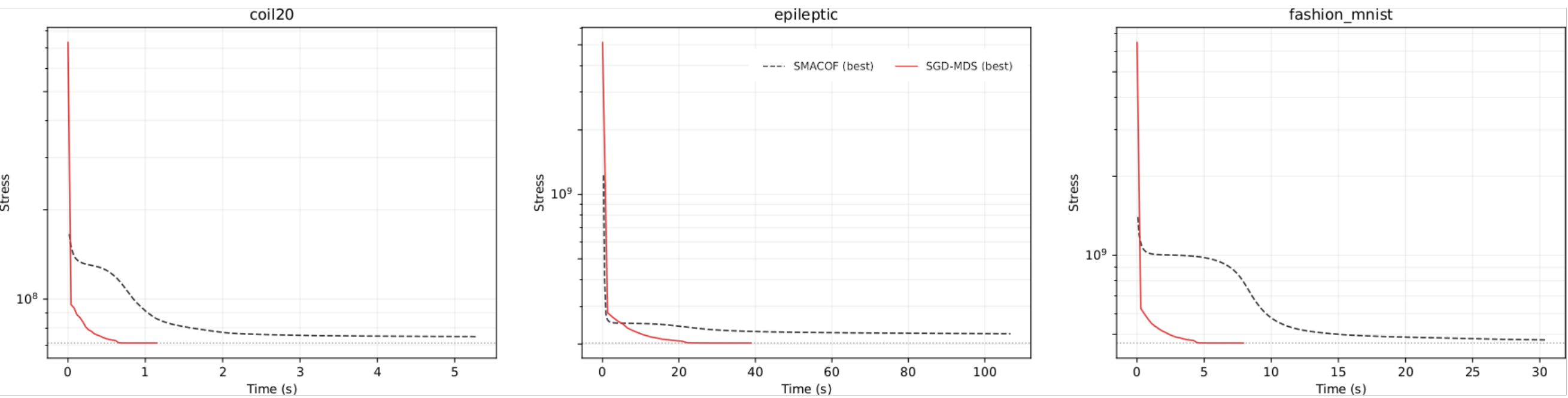}
    \includegraphics[width=0.9\linewidth]{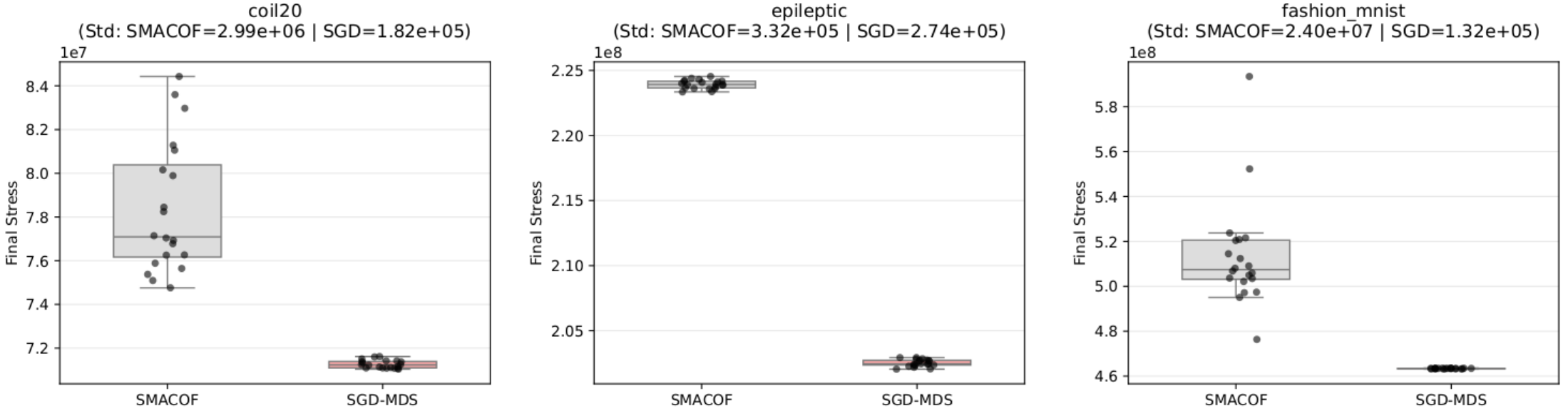}
    \caption{Results for 3 out of the 18 datasets in our benchmark on convergence time (top) and final quality (bottom).}
    \label{fig:results}
    \vspace{-0.5cm}

\end{figure*}

\section{Multi-dimensional Scaling by Stochastic Gradient Descent}
Our algorithm is based heavily on the SGD scheme from Zheng et al.~\cite{zheng2018graph}, with modifications for consideration of non-graph data. We describe coarse details here, and note the code is \href{https://github.com/DanielHangan01/scikit-learn}{available}.

The key observation is that each term in the stress function (Equation~\ref{eq:stress}) corresponds to a single pair $(i,j)$. Rather than evaluating the full objective, we sample one pair at a time and compute the gradient of its local contribution. The embedding is updated immediately after each such evaluation. This immediate update mechanism enables rapid early-stage reconfiguration of point positions, which we find empirically to accelerate convergence.

\paragraph{Stochastic Update Rule:}
In each step, the algorithm samples a pair $(i, j)$ and computes the gradient of the local stress contribution. The update direction is colinear with the displacement vector between the two points. 
Defining the normalized displacement vector $\mathbf{v}_{ij} = (x_i - x_j) / d_{ij}$, the update rule for point $x_i$ with learning rate $\eta(t)$ (absorbing all constants) is given by:
\begin{equation}
x_i \leftarrow x_i - \eta(t) \cdot w_{ij} \left( d_{ij} - \delta_{ij} \right) \mathbf{v}_{ij}
\end{equation}
The term $(d_{ij} - \delta_{ij})$ represents the magnitude of the violation of the metric constraint. Point $x_j$ is updated symmetrically in the opposite direction ($x_j \leftarrow x_j + \dots$). Note that this constitutes a simple stochastic gradient descent of the stress function.

\paragraph{Algorithmic Properties:}
While 
simple, the 
success of this approach relies on three properties that distinguish it from SMACOF:
\begin{compactitem}

    \item \textbf{Immediate Propagation:} Unlike SMACOF, which accumulates a global displacement vector before changes, SGD-MDS updates positions $x_i$ and $x_j$ in-place immediately. When point $i$ is selected again later in the same epoch (paired with point $k$), the new position of $x_i$ is used.
    This allows structural information to propagate across the projection rapidly within a single pass.

    \item \textbf{Stochastic Shuffling:} To prevent systematic update bias,
    the processing order of pairs is randomly permuted at the start of every epoch. 
    Random shuffling 
    ensures that, in expectation, the updates approximate the full gradient. This promotes broader exploration of the non-convex stress landscape.

    \item \textbf{Stability via Magnitude Clipping:} 
    Large dissimilarities can induce 
    large gradient steps, leading to numerical instability. Following Zheng et al.~\cite{zheng2018graph}, we cap the displacement magnitude of each update. 
    A pair is never moved beyond the configuration where $d_{ij}=\delta_{ij}$, ensuring that an update cannot overshoot the zero-stress condition. This preserves local structure while still allowing distant pairs to exert influence.

\end{compactitem}

\paragraph*{Implementation Details:}
We provide a fully \texttt{scikit-learn} compatible estimator, \texttt{SGD-MDS}, which inherits from \texttt{BaseEstimator}.
The core optimization loop is implemented in Cython.
For smaller datasets ($N \lesssim 20,000$), the \textbf{Precomputed Mode} prioritizes computational speed, by first precalculating the distances between all pairs of points.
Before each epoch, the pairs are shuffled and iterated over to ensure sampling without replacement. A full iteration over all pairs constitutes a single epoch.

For larger datasets ($N > 20,000$), where storing an $N \times N$ float matrix is infeasible, we introduce a \textbf{Lazy Mode}. This strategy bypasses the creation of the pairs list. 
Instead, it samples indices $(i,j)$ with replacement, calculating dissimilarity on-the-fly.
It achieves $O(1)$ auxiliary memory usage, at the cost of more time per epoch.
We employ an automatic learning rate scaling mechanism that normalizes the initial rate $\eta_0 \propto 1 / w_{\min}$, ensuring stable step sizes regardless of input scale. { Based on a grid search over $\eta_0 \in \{0.1, 0.5, 1.0\}$, we chose $\eta_0 = 0.5$ to balance convergence speed with the instability seen at higher rates.} 
Furthermore, we utilize a hybrid decay schedule inspired by the ``convergent schedule'' of Zheng et al.~\cite{zheng2018graph}.
We combine an \emph{exponential decay} phase 
with a \emph{harmonic decay} ($\eta(t) \propto 1/t$). 
The original formulation determines the phase transition dynamically based on the weight ratio $w_{\max}/w_{\min}$ but for 
standard 
MDS applications, weights are typically uniform, and this ratio is less useful.
Our scheduler enforces a fixed transition point at 40\% of epochs ensuring large steps early on and convergence later in the optimization. 
\begin{figure*}[t]
    \centering
    \includegraphics[height=4cm]{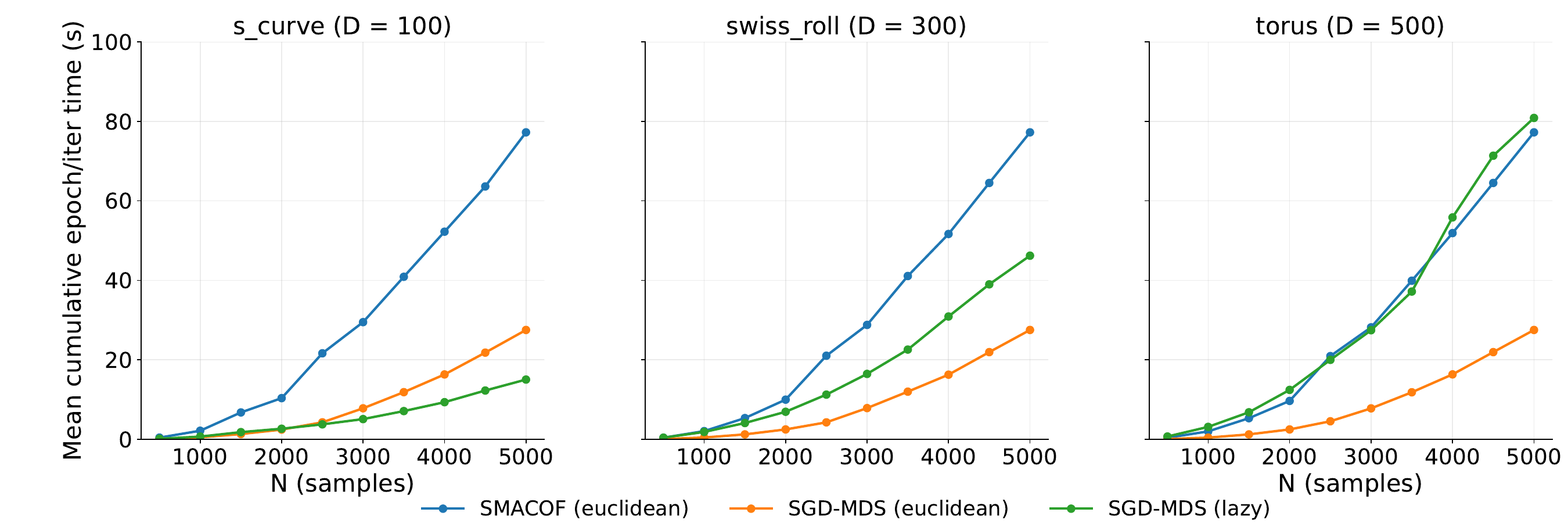}
    \caption{The runtime as $N$ increases for three synthetic datasets.}
    \label{fig:runtime}
    \vspace{-0.5cm}
\end{figure*}

\section{Evaluation}

We evaluate the performance of \texttt{SGD-MDS} against the standard \texttt{scikit-learn} implementation of SMACOF using the dataset collection found in Espadoto et al.'s survey~\cite{espadoto2019toward}.  
Experiments were conducted on commodity hardware\footnote{Intel i7 (20 cores), 16 GB RAM, running Pop!\_OS 22.04.}, comparing convergence speed and final stress quality. All data is collected over multiple random initializations per dataset.

\paragraph*{Convergence and Solution Quality:}
Overall, our experiments demonstrate that the stochastic approach consistently achieves rapid convergence, similar to the GD setting. Across all datasets, the SGD solver approaches a stable configuration within 15 to 30 epochs, whereas SMACOF typically requires hundreds of iterations to reach comparable termination criteria. In terms of solution quality (stress), \texttt{SGD-MDS} performs well, achieving consistently better stress in 14 out of the 18 instances in the benchmark. We summarize our findings, and provide a subset of results;
see Figure~\ref{fig:results}. Further detail can be found in supplemental material.

In contrast to SMACOF's sensitivity to initial configuration~\cite{borg2005modern}, SGD-MDS is generally consistent in finding similar minimums.
We observe 4 instances in which the final stress quality is not definitively more consistent than SMACOF. A notable example is the IMDB dataset, where SMACOF achieved a $2.5\%$ lower stress in less time.  We note that this dataset is particularly difficult for distance-based embeddings, as it is an example of data with high norm concentration.
In this case, the added noise induced by the stochastic optimization may hinder stable convergence.

\paragraph*{Computational Efficiency:}

In its standard (precomputed) mode, \texttt{SGD-MDS} demonstrates a significant advantage in total time to convergence. 
This is demonstrated in Figure~\ref{fig:runtime}, where we generate synthetic data of increasing size and measure time to convergence for SMACOF, Precomputed SGD-MDS, and lazy SGD-MDS. In all instances, we see that the total time for the precomputed mode is already twice as fast as SMACOF at $N=3000$, with differences becoming larger for larger $N$. The speedup from the lazy mode quickly diminishes as $D$ increases.
We note the trade-off in operation costs: a single epoch of SGD is computationally heavier than a single SMACOF iteration because it cannot leverage highly optimized matrix operations. However, this higher per-pass cost is outweighed by SGD's rapid convergence. 

\section{Limitations and Conclusion}

The primary limitation of our current implementation is its inability to leverage linear algebra operations.
The SGD approach is inherently sequential; it processes pairs one-at-a-time. { Our benchmark includes limited datasets, and scalability on datasets of larger sizes becomes a bottleneck. Further performance gains, such as employing stress approximation schemes used in graph drawing, are future work.}
The SGD approach offers no theoretical guarantees, unlike SMACOF. 
Finally, more rigorous experiments on a larger variety of datasets are needed to fully understand the benefits and tradeoffs of SGD-MDS over SMACOF.

In this work, we have shown of an example of a critical result found in graph drawing that transfers with little changes to dimensionality reduction.
While these fields have traditionally relied on distinct optimization paradigms, both fields can benefit through collaboration and understanding of the other's results. 
\bibliographystyle{eg-alpha-doi}  
\bibliography{references}

\end{document}